\pdfoutput=1
\documentclass[10pt,twocolumn,letterpaper]{article}
\usepackage{iccv}
\usepackage{times}
\usepackage{epsfig}
\usepackage{graphicx}
\usepackage{amsmath}
\usepackage{amssymb}
\usepackage{animate}
\usepackage{float}
\usepackage{colortbl}
\usepackage{enumerate}
\definecolor{lightgray}{rgb}{0.83, 0.83, 0.83}
\usepackage[pagebackref=true,breaklinks=true,letterpaper=true,colorlinks,bookmarks=false]{hyperref}

\iccvfinalcopy % *** Uncomment this line for the final submission

\def\iccvPaperID{2945} % *** Enter the ICCV Paper ID here

\ificcvfinal\fi

\begin{document}

%%%%%%%%% TITLE
% \title{VideoGen: Towards High Quality Video Generation with Cascaded 
% \newline Reference-Based Latent Diffusion Models}
% \title{Create-A-Video: A Reference-guided Latent Diffusion Approach to Text-to-Video}
% \title{Create-A-Video: Reference-guided Latent Diffusion for Text-to-Video}
% \title{Generate-A-Video: A Reference-Guided Latent Diffusion Approach for 
% High Definition Text-to-Video Generation}
\title{VideoGen: A Reference-Guided Latent Diffusion Approach for 
High Definition Text-to-Video Generation}
%\title{VideoGen: Reference-Based Text-to-Video Generation with 
%\newline Cascaded Latent Diffusion Models}

\author{Xin Li, Wenqing Chu, Ye Wu, Weihang Yuan, Fanglong Liu, Qi Zhang, \\
Fu Li, Haocheng Feng, Errui Ding, Jingdong Wang\\
Department of Computer Vision Technology (VIS), Baidu Inc.\\
{\tt\small lixin41@baidu.com}
}

\maketitle
% \input{VideoGen/figs/demo.tex}

% Remove page # from the first page of camera-ready.
\ificcvfinal\thispagestyle{empty}\fi
%%%%%%%%% ABSTRACT
\begin{abstract}
 In this paper, we present VideoGen,
a text-to-video generation approach,
which can generate a high-definition video with high frame fidelity and strong temporal consistency
using reference-guided latent diffusion.
We leverage an off-the-shelf text-to-image generation model,
e.g., Stable Diffusion,
to generate an image
with high content quality
from the text prompt,
as a reference image to guide video generation.
Then,
we introduce an efficient
cascaded latent diffusion module
conditioned on both the reference image
and the text prompt,
for generating latent video representations,
followed by
a flow-based temporal upsampling step
to improve the temporal resolution.
Finally,
we map latent video representations
into a high-definition video
through an enhanced video decoder.
During training,
we use the first frame of a ground-truth video
as the reference image for training the cascaded latent diffusion module.
The main characterises of our approach include:
the reference image generated by the text-to-image model improves the visual fidelity;
using it as the condition makes the diffusion model focus more on learning the video dynamics;
and the video decoder is trained over unlabeled video data, thus benefiting from high-quality easily-available videos. 
VideoGen sets a new state-of-the-art 
in text-to-video generation
in terms of both qualitative and quantitative evaluation. 
See \url{https://videogen.github.io/VideoGen/} for more samples.

   % Text-to-video generation has recently shown promising progress. While this is still a challenging task due to uncontrollable content, temporal consistency, computational complexity and video quality, etc. In this paper, we propose VideoGen, a text-to-video generation method conditioned on a text prompt and a reference image, using cascaded diffusion models processed in the latent space. In particular, we use a reference image generated by StableDiffusion as an additional condition, which makes the task less difficult and the generated videos are of higher quality and more controllable. The whole generation is proceeded in a cascade manner to facilitate the model training.  
   %
   % In addition, we propose a flow-guided video latent interpolation diffusion model, compared with pixel-level interpolation, which can produce arbitrary multipliers of interpolation frames without suffering from blurring. Finally, an enhanced video-level decoder is used to make the generated video more clearer and smoother. Since models are trained in the latent space, we reduce the resources required for training while still guaranteeing high-quality video generation. Extensive experiments demonstrate the superiority of our method. See http://xxx for more samples.
\end{abstract}
%%%%%%%%% BODY TEXT
\section{Introduction}
\begin{figure*}[t]
	\centering	\includegraphics[width=1.0\textwidth]{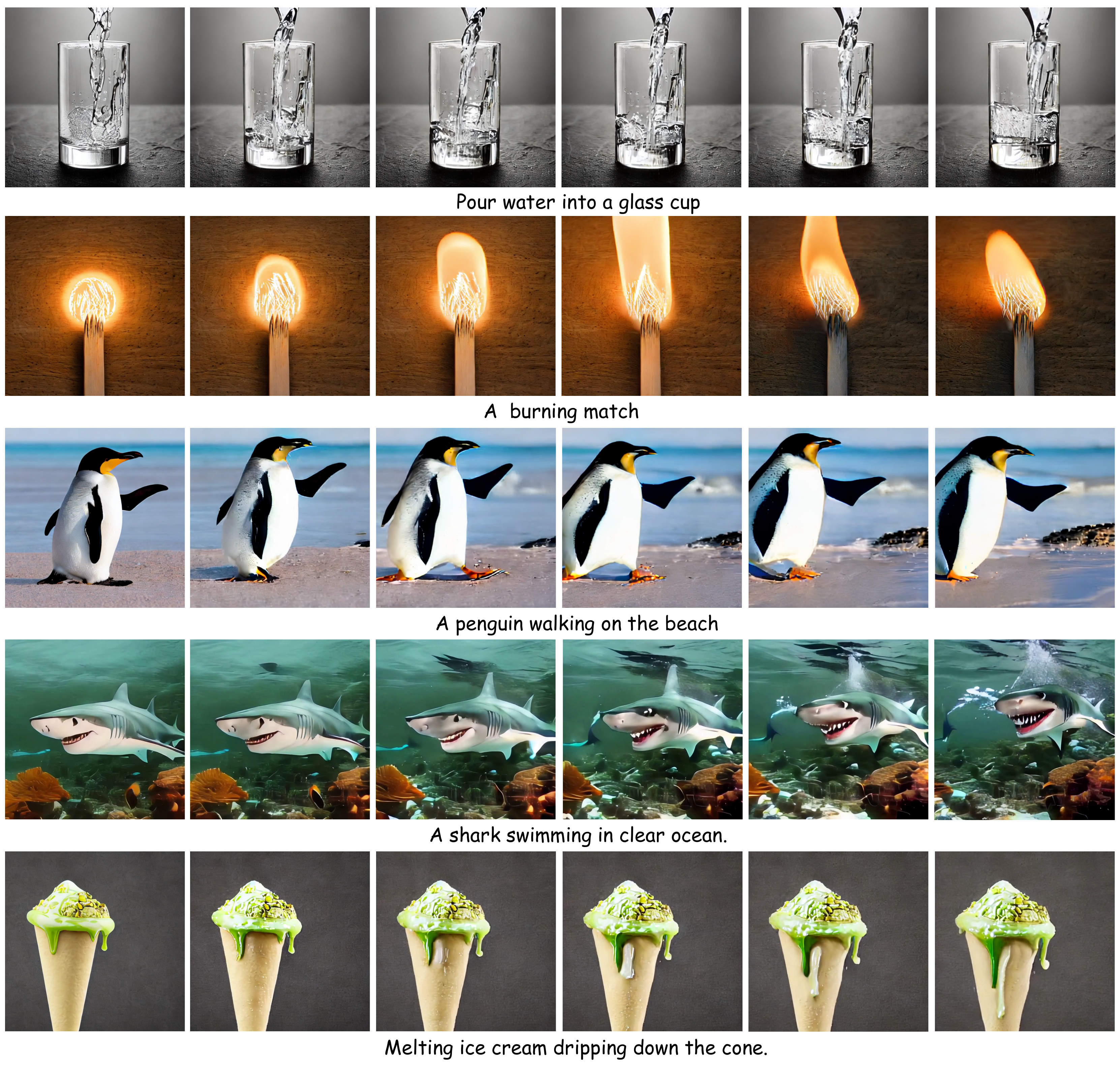}
	\caption{T2V generation examples of VideoGen. Our generated videos have rich texture details and stable temporal consistency. It is strongly recommended to zoom in to see more details.}
 %
 %Videos and more samples are available at \url{https://videogen.github.io/VideoGen/}
	\label{fig:frameresults}
\end{figure*}

There have been great progress
in text-to-image (\texttt{T2I}) generation systems,
such as DALL-E2~\cite{esser2023structure},
Imagen~\cite{saharia2022photorealistic},
Cogview~\cite{ding2021cogview},
%diffusion models~\cite{ho2020denoising},
Latent Diffusion~\cite{rombach2022high},
and so on.
In contrast,
text-to-video (\texttt{T2V}) generation,
creating videos from text description, 
is still a challenging task 
as it requires not only high-quality visual content, but also temporally-smooth and realistic motion that matches the text. 
Moreover, it is hard to find large-scale datasets of text-video pairs.

In addition to 
extending 
the \texttt{T2I} network architecture,
several recent \texttt{T2V} techniques 
explore the trained \texttt{T2I} model
for improving the visual fidelity,
e.g., 
utilizing the \texttt{T2I} model weights,
or exploring image-text data.
For example,
CogVideo~\cite{hong2022cogvideo} and Make-A-Video~\cite{singer2022make}
make use of the \texttt{T2I} model,
by freezing or fine-tuning the \texttt{T2I} model weights.
N\"UWA~\cite{wu2022nuwa} and Imagen Video~\cite{ho2022imagen}
instead explore image-text pairs
to improve \texttt{T2V} model training,
through pre-training or joint-training.

In this paper,
we propose VideoGen for 
generating a high-quality and temporally-smooth video
from a text description.
We leverage a \texttt{T2I} model
to generate a high-quality image,
which is used as a reference
to guide \texttt{T2V} generation.
Then,
we adopt a cascaded latent video diffusion module,
conditioned on the reference image and the text description,
to generate a sequence of high-resolution smooth latent representations.
We optionally use a flow-based scheme
to temporally upsample the latent representation sequence.
Finally,
we learn a video decoder to
map the latent representation sequence
to a video.

% \color[red]{we train the model???}

The benefits of using a \texttt{T2I} model to generate a reference image lie in two-fold.
On the one hand,
the visual fidelity of the generated video
is increased.
This benefits from that 
our approach makes use of the large dataset
of image-text pairs,
which is richer and more diverse than the dataset
of video-text pairs,
through using the \texttt{T2I} model.
This is more training-efficient compared
to Imagen Video that
needs to use the image-text pairs for joint training.
On the other hand,
using the reference image
to guide the cascaded latent video diffusion model
frees the diffusion model from learning visual content,
and makes it focus more on learning the video dynamics.
We believe that
this is an extra advantage 
compared to the methods
merely using the \texttt{T2I} model parameters~\cite{hong2022cogvideo, singer2022make}.

Furthermore,
our video decoder
only needs 
the latent representation sequence as input
to generate a video,
without requiring the text description.
This enables us to train the video decoder 
over a larger set of
easily-available unlabeled (unpaired) videos
other than only video-text pairs.
As a result,
our approach benefits 
from high-quality video data,
improving motion smoothness and motion realism
of the generated video. 
Our key contributions are as follows:
\begin{itemize}
\item We leverage an off-the-shelf \texttt{T2I} model 
to generate an image from text description
as a reference image,
for improving frame content quality.

\item We present an efficient and effective cascaded latent video diffusion model 
conditioned on the text description,
as well as the reference image as the condition
which makes the diffusion model focus more
on learning the video motion.

\item We are able to train the video decoder
using easily-available unlabeled (unpaired) 
high-quality video data,
which boosts visual fidelity and 
motion consistency
of the generated video.

\item We evaluate VideoGen against 
representative \texttt{T2V} methods
and present state-of-the-art results
in terms of
quantitative and qualitative measures.

% and text description 
% apropose a flow-guided video latent interpolation diffusion model and an enhanced video-level decoder to make the generated videos smoother and clearer.
% \item VideoGen uses multiple diffusion models cascaded in the latent space to greatly reduce the resources required for training, and extensive experiments demonstrate the superiority of our method. 
\end{itemize}

% In addition to high-quality visual content,

% raises an extra requirement,
% temporally-smooth and realistic motion that matches the text.
% Furthermore, large scale data of video-text pairs is not available
% as easily as image-text pairs.
% Thus, \texttt{T2V} generation thus remains a challenge.

\begin{figure*}[t]
	\centering
	\includegraphics[width=0.99\textwidth]{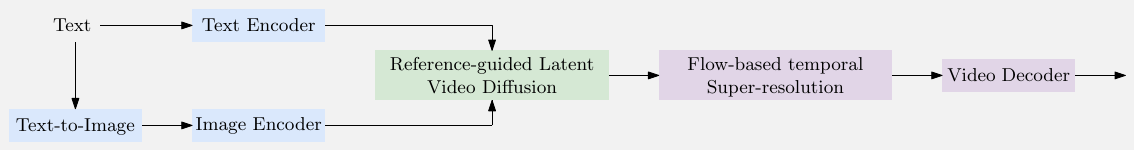}
 	\caption{The VideoGen inference pipeline.
The input text is fed into a pretrained Text-to-Image generation model,
generating a reference image.
The reference image and the input text
are sent to a pretrained Image Encoder
and a pretained Text Encoder.
The output text and image embeddings
are used as the conditions
of Reference-guided Latent Video Diffusion,
outputting the latent video representation. 
Then Flow-based temporal Super-resolution increases
the temporal resolution,
and is followed by Video Decoder,
generating the final video.
During the training process,
the reference image is the first frame 
of the video. }
	% \caption{Inference pipeline of VideoGen. Given a text prompt, we leverage a "T2I" model (e.g., Stable Diffusion) to generate a reference image, which along with text prompt, are used as conditions for a reference-based video latent generation (RVLG) diffusion models. Subsequently, we use two conditioned multi-scale video latent super-resolution (CMSR) diffusion models to gradually increase the spatial resolution. Afterwards, we propose a flow-guided video latent interpolation (FVLI) to interpolate the video latent code to a desired duration. Finally, an enhanced video-level decoder (EVD) is used to convert the video latent code to the generated video, which avoids temporal flickering and makes the generated video clearer.}
	\label{fig:pipeline}
\end{figure*}

\section{Related Work}
% \subsection{Text-to-Image Generation with DMs}
\noindent\textbf{Diffusion models.}
The generative technology has experienced rapid development, from the generative adversarial networks \cite{goodfellow2020generative} in the past few years to the very popular diffusion models recently.
Diffusion models \cite{sohl2015deep, ho2020denoising} have shown surprising potential and made great progress in generative tasks, such as text-to-speech \cite{chen2020wavegrad, chen2021wavegrad, kong2020diffwave}, text-to-image \cite{saharia2022photorealistic, ramesh2022hierarchical, peebles2022scalable, rombach2022high, nichol2021glide, balaji2022ediffi, gafni2022make, chang2023muse}, text-to-3D \cite{poole2022dreamfusion, watson2022novel}, text-to-video \cite{ho2022video, singer2022make, he2022latent, zhou2022magicvideo, ho2022imagen, wu2022tune, hong2022cogvideo}, image2image \cite{saharia2022image, brooks2022instructpix2pix, valevski2022unitune, zhang2022sine, saharia2022palette, bar2022text2live} and vid2vid \cite{esser2023structure, bar2022text2live}. 
Especially in the generation of images, such as Stable Diffusion \cite{rombach2022high}, has reached the level of professional illustrators, which greatly improves the work efficiency of artists. 
%Most recently, a variety of applications based on Stable Diffusion have emerged in an endless stream, making people see the huge business prospects of AIGC. 

\noindent\textbf{Text-to-image generation.}
The past years have witnessed tremendous
progress in image-to-text generation.
The early systems are mainly based on GAN~\cite{goodfellow2020generative},
e.g., StyleCLIP ~\cite{patashnik2021styleclip}, StyleGAN-NADA~\cite{gal2022stylegan}, VQGAN-CLIP~\cite{crowson2022vqgan}, StyleT2I ~\cite{li2022stylet2i}. 
The most recent success  
is from the development of 
denoising diffusion model~\cite{ho2020denoising}
and its efficient extension,
latent diffusion model~\cite{rombach2022high}.
Examples include: 
DALL-E \cite{ramesh2021zero},
DALL-E2 \cite{ramesh2022hierarchical},
Imagen \cite{saharia2022photorealistic},
Stable Diffusion \cite{rombach2022high},
CogView~\cite{ding2021cogview},
Parti~\cite{yu2022scaling}, GLIDE~\cite{nichol2021glide}.

Our approach takes advantages of latent diffusion model ~\cite{rombach2022high} for text-to-video generation.
This not only improves the diffusion sampling efficiency,
but also allows 
to design the video decoder that only relies on videos, not on texts,
allowing that the video decoder can be trained
on high-quality unlabeled videos.

% DDPM  ushers in a new era of image synthesis with diffusion models (DMs). The pioneering DALL-E \cite{ramesh2021zero} opened eyes to the possibility of generating images from text. 
% Later Imagen \cite{saharia2022photorealistic}, DALLE-2 \cite{ramesh2022hierarchical} and StableDiffusion \cite{rombach2022high}, etc. all showed the strong capabilities in image synthesis tasks, greatly surpassing previous GAN-based methods in terms of quality and diversity. 
% Especially the open source LDM-based StableDiffusion, the marvelous generated images make it very popular.
% LDM \cite{rombach2022high} proposes a latent-based DM that drastically reduces the resources required for training compared to pixel-based DMs, while generating images that are competitive in clarity and resolution.
% In this paper, we also use the latent-based DMs in a cascading manner, which reduces the amount of calculation while ensuring the quality of the generated videos.
\noindent\textbf{Text-to-video generation.}
Early text-to-video techniques include:
leveraging
a VAE with recurrent attention,
e.g.,Sync-DRAW~\cite{mittal2017sync},
and extending GAN from image generation
to video generation~\cite{pan2017create,li2018video}.
Other developments include
GODIVA~\cite{wu2021godiva},
N\"UWA~\cite{wu2022nuwa},
CogVideo~\cite{hong2022cogvideo}.

More recent approaches include:
Tune-A-Video \cite{wu2022tune} and Dreamix \cite{molad2023dreamix} for applications with fine-tuning,
Make-A-Video~\cite{singer2022make}, MagicVideo~\cite{zhou2022magicvideo},
Video Diffusion Model~\cite{ho2022video} and 
Imagen Video~\cite{ho2022imagen}, 
latent video diffusion models~\cite{he2022latent},
which extend diffusion models
from image generation
to video generation,

Our approach differs from previous works in several aspects.
First,
our approach leverages the pretrained text-to-image
generation model
to generate a high-quality image
for guiding video generation,
leading to high visual fidelity
of the generated video.
This is clearly different from previous approaches. 
In Make-A-Video~\cite{singer2022make}, an image is used to generate an embedding to replace the text embedding for image animation. In contrast, our approach uses an image as reference to guide video content generation. What's more, the image in Make-A-Video is mapped to an embedding through CLIP image encoder, that is mainly about semantic. In contrast, our approach uses the encoder trained with auto-encoder, and the output latent contains both semantics and details for reconstruction. This is why the results of Make-A-Video are more blurry.
Second,
we adopt latent video diffusion model, 
leading to more efficient diffusion sampling
in comparison to  
Make-A-Video~\cite{singer2022make} and 
Imagen Video~\cite{ho2022imagen}.
Reference-guidance for latent video diffusion model makes our approach differ 
from~\cite{he2022latent} 
that only conducts the study on a small dataset.
Last, our design allows us to train the video decoder using high-quality unpaired videos.

\section{Approach}

Our approach VideoGen
receives a text description,
and generates a video.
The inference pipeline is depicted in  Figure~\ref{fig:pipeline}.
We generate a reference image 
from a pretrained and frozen Text-to-Image generation model.
We then compute the embeddings of
the input text and the reference image
from pretrained and frozen text and image encoders.
We send the two embeddings
as the conditions for reference-guided latent video diffusion
for generating latent video representation,
followed by a 
flow-based temporal super-resolution
module.
Finally, we map the latent video representation
to a video through a video decoder.

\subsection{Reference Image Generation}
We leverage an off-the-shelf 
text-to-image (\texttt{\texttt{T2I}}) generation model,
which is trained over a large
set of image-text pairs
and can generate high-quality image.
In our implementation,
we adopt the SOTA model, 
Stable Diffusion\footnote{https://github.com/CompVis/stable-diffusion} 
without any processing.
We feed the text prompt 
into the \texttt{\texttt{T2I}} model.
The resulting high-fidelity image is used 
as a reference image,
and plays a critical role
for effectively guiding 
subsequent
latent representation sequence generation.
During the training,
we simply pick the first frame of the video
as the reference,
which empirically works well.

\subsection{Reference-Guided Latent Video Diffusion}
Cascaded latent video diffusion
consists of three consecutive components:
a latent video representation diffusion network,
generating representations of spatial resolution $16 \times 16$
and temporal resolution $16$,
and two spatially super-resolution diffusion networks, raising the spatial resolutions
to $32 \times 32$ and $64 \times 64$.

\noindent\textbf{Architecture.}
We extend the $2$D latent diffusion model~\cite{rombach2022high}
to the $3$D latent diffusion model
through taking into consideration the temporal dimension.
We make two main modifications
over the key building block
that now supports both spatial and temporal dimensions.

Following Make-A-Video~\cite{singer2022make},
we simply stack a $1$D temporal convolution
following
each $2$D spatial convolutional layer
in the network.
The $2$D spatial convolution
is conducted for each frame separately,
e.g., $16$ frames in our implementation.
Similarly, the $1$D temporal convolution 
is conducted for each spatial position separately,
e.g., $16 \times 16$,
$32 \times 32$,
and $64 \times 64$
for the three diffusion networks.
Similar to Make-A-Video~\cite{singer2022make}.
such a modification to the building block
enables us to use the pretrained $\texttt{T2I}$ model parameters to initialize
the $2$D convolutions.
Similarly,
we stack a temporal attention
following each spatial attention.

% The three diffusion networks use the same architecture.
% The first network use the noise as the input
% and the output representation 
% is spatially upsampled simply with a bilinear interpolation 
% as the input of the second diffusion network.
% The output of the second diffusion network
% is similarly upsampled
% as the input of the third diffusion network.

\noindent\textbf{Condition injection.}
We follow the scheme in LDM~\cite{rombach2022high}
to inject the text embedding
into the network 
using cross-attention.
We project the text description
into an intermediate representation 
through a pretrained text encoder,
CLIP text encoder in our implementation.
The intermediate representation
is then mapped into each diffusion network
using a cross-attention layer.

The later diffusion network
uses the bilinear 
$2\times$ upsampled
representation output from the last diffusion network
as an extra condition
and concatenates it
into the input.
We follow Make-A-Video~\cite{singer2022make}
to use FPS as a condition
and inject its embedding into each diffusion model.

We project the reference image
to a representation
through a pretrained image encoder.
In our implementation,
we use the image encoder of the auto-encoder in Stable Diffusion,
and process the image with three resolutions
($16 \times 16$,
$32 \times 32$,
and $64 \times 64$),
each corresponding to a diffusion network.
We inject the representation of the reference image
into the network
by concatenating it 
with the first-frame representation 
of the input of the diffusion model,
and concatenating zero representations
with the representations corresponding to other frames.

\begin{figure*}[t]
	\centering	\includegraphics[width=1.0\textwidth]{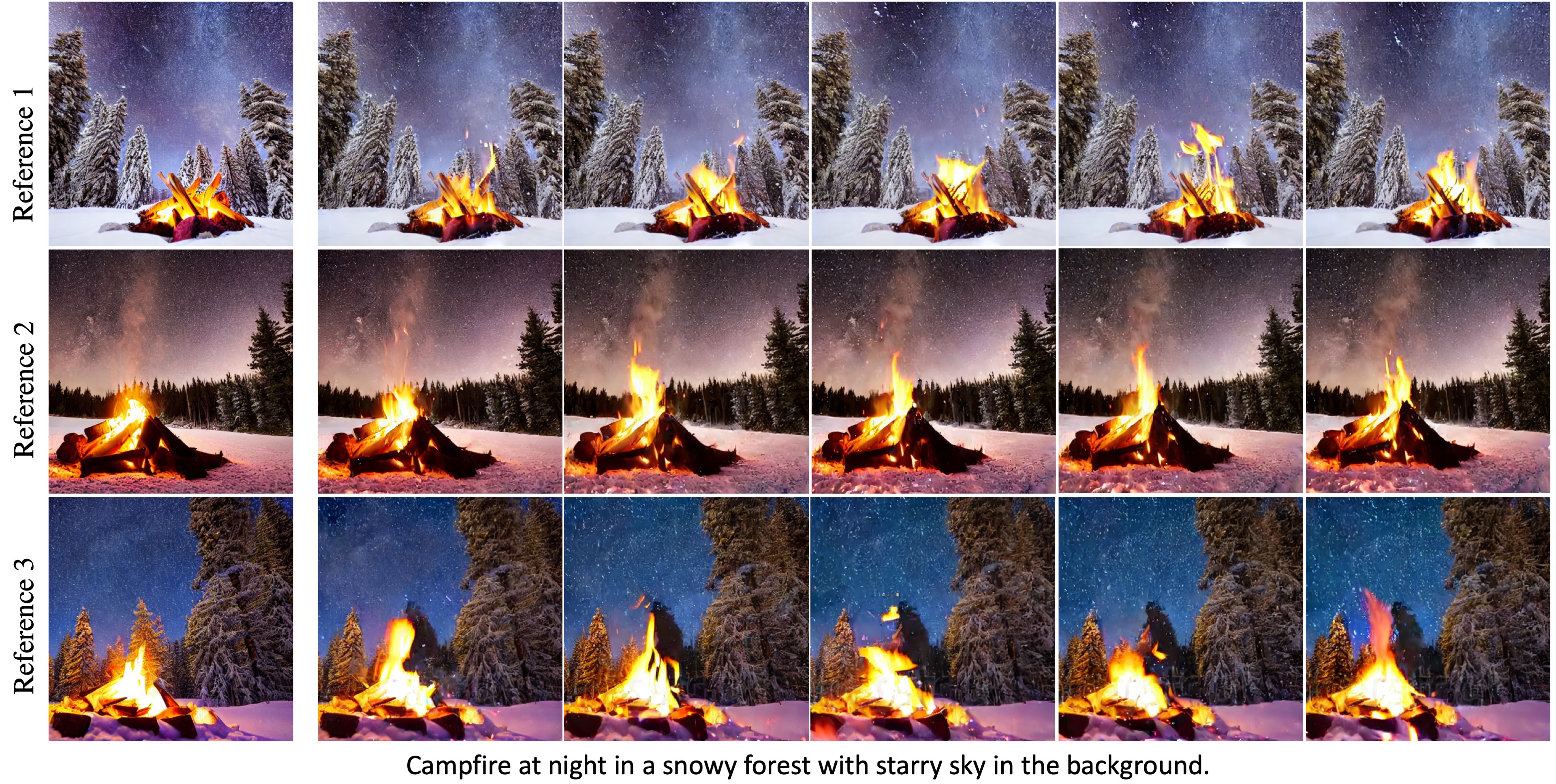}
	\caption{For a text prompt, different reference images generate different videos.}
	\label{fig:dif_ref}
\end{figure*}

\subsection{Flow-based Temporal Super-resolution}
We perform temporal super-resolution
in the latent representation space.
We estimate the motion flow
according to the representations
using a latent motion flow network.
Then we warp the representations
according to the estimated motion flow,
and obtain a coarse longer video representations
with $2\times$ upsampling.
We next send each warped representation
to a denoising diffusion network as a condition
to get a refined representation.
The final warp representation is a combination 
of the low-frequency component of the warped coarse representation
and the high-frequency component of the refined representation.
Consistent to the observation~\cite{choi2021ilvr},
our experiments find that 
the combined representations lead to 
more stable video generation.
We perform this process three times
and get $8\times$ upsampled video representations.

\subsection{Video Decoder}
The video decoder
maps the video from the latent representation space
to pixel space.
We modify the Stable Diffusion $8\times$ upsampling image decoder
for the video decoder.
We stack a $1$D temporal convolution following
each $2$D convolution
and a temporal attention following each spatial attention.
This modification also allows us
to initialize the parameters
of $2$D convolutions and spatial attentions
in the video decoder
using the parameters of the pretrained image decoder.

\subsection{Training}
Our approach leverages
existing models,
e.g., CLIP text encoder for text description encoding,
Stable Diffusion \texttt{T2I} generation model
for reference image generation,
Stable Diffusion image encoder for reference image encoding.
In our implementation,
we freeze the three models without retraining.
The other three modules are independently trained from the video data
with the help of pretrained image models.
The details are as follows.

\noindent\textbf{Reference-guided cascaded
latent video diffusion.}
We compute the video representations
by sending each frame into the image encoder
as the denoising diffusion target.
At each stage, the video spatial resolution
is processed
to match the spatial resolution of the latent representations.
We simply pick the first frame in the video
as the reference image for training.

The $2$D convolution and spatial attention parameters of the first diffusion network
are initialized from the pretrained Stable Diffusion \texttt{T2I} generation model.
The temporal convolution and attention layers are initialized as the identity function.
The second (third) diffusion network is
initialized as the weights
of the trained first (second) diffusion network.
The three diffusion networks
are only the components receiving
video-text pairs,
WebVid-10M \cite{bain2021frozen},
for training.

\noindent\textbf{Flow-based temporal super-resolution.}
We estimate the motion flow 
by extending IFRNet~\cite{kong2022ifrnet}
from the pixel space
to the latent representation space.
We slightly modify the IFRNet architecture
and simply change the first layer
for processing latent representations.
The ground-truth motion flow
in the latent representation space 
is computed as:
compute the motion flow in the pixel space using the pretrained IFRNet
and resize the motion flow to the spatial size of the latent representation space.

The input representations
of the flow-based temporal super-resolution part 
are directly computed 
from low temporal-resolution video.
The ground-truth target representations
of the denoising diffusion network for
warped representation refinement
are constructed 
by feeding the frames
of high FPS video
into the image encoder. 

\noindent\textbf{Video decoder.}
The $2$D convolution and spatial attention weights are initialized from the pretrained Stable Diffusion image decoder, and the temporal convolution and attention are initialized as the identify function.
During the training, we use the image encoder in StableDiffusion to extract video latent representations.
We apply degradations (adding noise, blurring, and compression), which are introduced in BSRGAN \cite{zhang2021designing},
to the video, and extract the latent representations.
The target video is still the original video,
and without any processing. 
Video decoder and flow-based temporal super-resolution network
are trained on
unpaired videos 
with 40$K$ clips of $100$ frames that are collected
from YouTube.

\begin{figure*}[t]
	\centering	\includegraphics[width=1.0\textwidth]{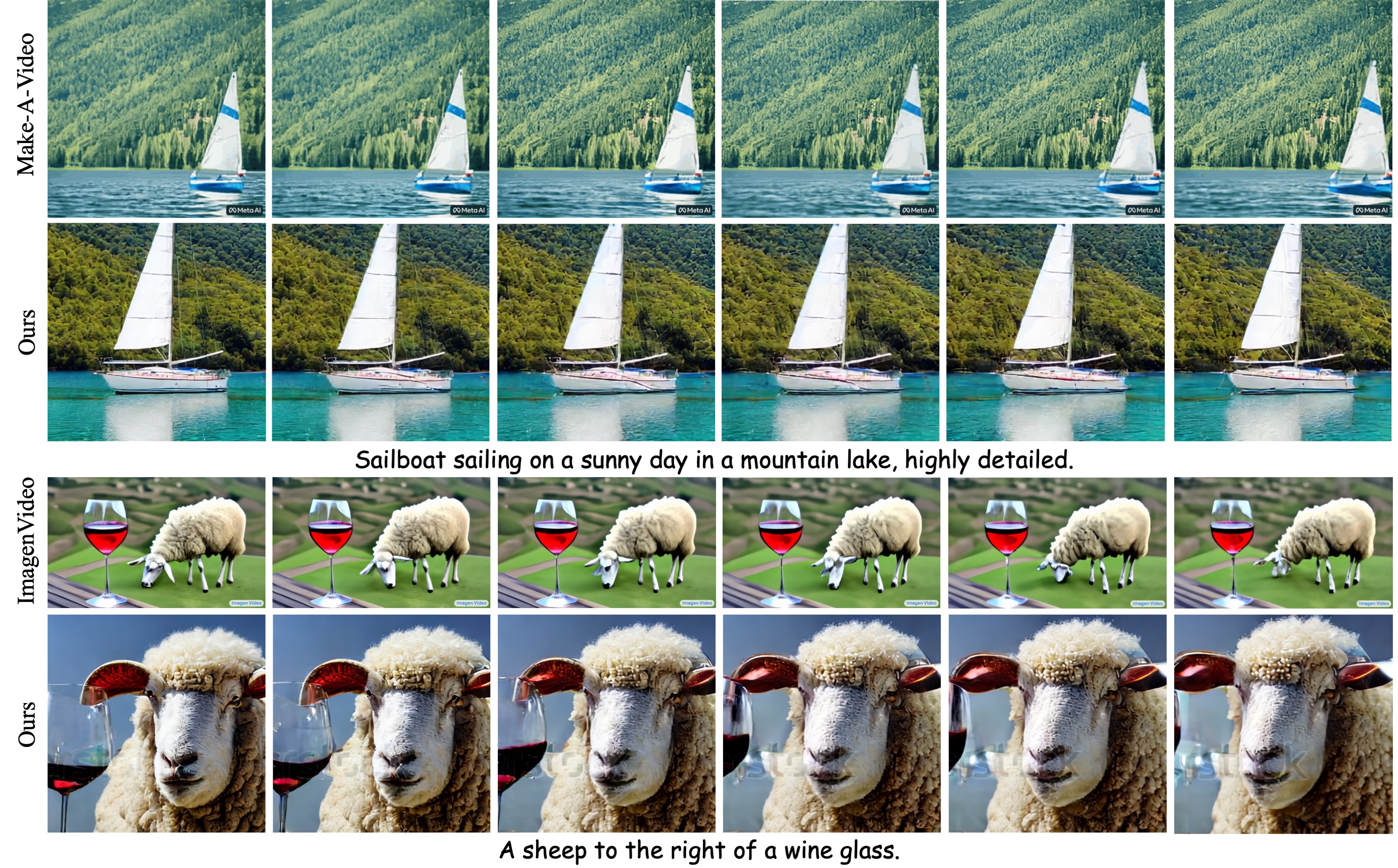}
	\caption{Qualitative comparison with Make-A-Video and Imagen Video.  Compared with Make-A-Video, the lake ripples, boats and trees in our video are clearer. Similarly, although the video resolution of Imagen Video reaches 1280$\times$768, the frames are very blurry compared with our result. 
 %Please zoom in to see clearer image details.
 The watermark in the last row is because the videos in the training set WebVid-10M contain the "shutterstock" watermark.}
	\label{fig:ab_qualitive_com}
\end{figure*}

\section{Experiments} \label{exp}
\begin{table*}[t]
\centering
%\resizebox{0.48\textwidth}{!}{
\caption{\texttt{T2V} results on UCF-101. We report the performance
for zero-shot and fine-tuning settings.}
\setlength{\tabcolsep}{22pt}
\renewcommand{\arraystretch}{1.3}
\footnotesize
\label{table:UCF101}
\begin{tabular}{l c c c c c}
\hline
Method & Pretrain & Class & Resolution & IS $\uparrow$ & FVD $\downarrow$ \\
\hline
  \multicolumn{6}{c}{\centering Zero-Shot Setting} \\
\hline
CogVideo (Chinese) & Yes & Yes & 480$\times$480 & 23.55 & 751.34 \\
CogVideo (English) & Yes & Yes & 480$\times$480 & 25.27 & 701.59 \\
Make-A-Video & Yes & Yes & 256$\times$256  & 33.00 & 367.23 \\
\rowcolor{lightgray} Ours & Yes & Yes & 256$\times$256 & 71.61 $\pm$ 0.24 & 554 $\pm$ 23\\
\hline
  \multicolumn{6}{c}{\centering Fine-tuning Setting} \\
\hline
TGANv2 & No & No & 128$\times$128 & 26.60 $\pm$ 0.47 & - \\
DIGAN & No & No & - & 32.70 $\pm$ 0.35 & 577 $\pm$ 22 \\
MoCoGAN-HD & No & No & 256$\times$256 & 33.95 $\pm$ 0.25 & 700 $\pm$ 24 \\
CogVideo & Yes & Yes & 160$\times$160 & 50.46 & 626 \\
VDM & No & No & 64$\times$64 & 57.80 $\pm$ 1.3 & - \\
LVDM & No & No & 256$\times$256 & - & 372 $\pm$ 11 \\
TATS-base & Yes & Yes & 128$\times$128 & 79.28 $\pm$ 0.38 & 278 $\pm$ 11 \\
Make-A-Video & Yes & Yes & 256$\times$256 &  82.55 & 81.25 \\
\rowcolor{lightgray} Ours & Yes & Yes & 256$\times$256 & 82.78 $\pm$ 0.34 & 345 $\pm$ 15 \\
\hline
\end{tabular}
% }
\end{table*}

\begin{table}[h]
\setlength{\tabcolsep}{8.5pt}
\renewcommand{\arraystretch}{1.3}
\footnotesize
\caption{\texttt{T2V} results on MSR-VTT. We report average CLIPSIM scores to evaluate the text-video alignment.}
\label{table:MSR-VTT}
\centering
% \resizebox{0.48\textwidth}{!}{
\begin{tabular}{l c c c}
\hline
Method & Zero-Shot & Resolution & CLIPSIM $\uparrow$  \\
\hline
GODIVA & No & 128$\times$128 & 0.2402 \\
N{\"u}wa & No  & 336$\times$336 & 0.2439 \\
CogVideo (Chinese) & Yes & 480$\times$480 & 0.2614\\
CogVideo (English) & Yes & 480$\times$480 & 0.2631 \\
Make-A-Video & Yes & 256$\times$256  & 0.3049 \\
\rowcolor{lightgray}Ours & Yes & 256$\times$256 & 0.3127 \\
\hline
\end{tabular}
% }
\end{table}
\subsection{Datasets and Metrics}
% \textbf{Implementation Details.} 
% Our three CRVLs generate 16-frame video latent codes with channel 4 and resolutions of 16$\times$16, 32$\times$32, and 64$\times$64, respectively.
% By using FVLI multiple times, we can achieve 2$\times$, 4$\times$, 8$\times$ temporal video latent interpolation.
% At last, with a 8$\times$ spatial upsampling EVD, we can generate videos with frame number between 16 and 128, and a resolution of 512*512.
% Of course, on this basis, we can generate videos with larger resolution through other existing video super-division methods.
% The three CRVLs, FVLI and EDV models in our VideoGen are all trained independently.
% We alternately train our CRVLs using text-image pairs from LAION-400M \cite{schuhmann2021laion} and text-video pairs from WebVid-10M \cite{bain2021frozen}.
% When it is the turn of text-to-image training, we bypass the temporal convolution and attention modules and only train the spatial parameters.
% We adopt a progressive training strategy, first training the lowest-resolution CRVL1, and then using the trained CRVL1 parameters to start CRVL2 training, finally CRVL3.
% As a comparison, we tried to train CRVL3 first, but it is difficult to converge because the resolution is large and the training batch size is too small.
% In the training phase, the reference image of CRVL uses the first frame of the training video clip, while in the inference phase, we use a T2I model StableDiffusion to generate the reference image.

%\noindent\textbf{Implementation details.}
%\vspace{1mm}
%\noindent\textbf{Training datasets.}
We adopt the publicly available dataset of video-text
pairs from WebVid-$10M$~\cite{bain2021frozen}
for training the reference-guided
cascaded latent video diffusion network.
We collected over $2,000$ $4K$-resolution videos
of $60$ FPS from YouTube and 
extracted 40000 clips 
for training the flow-based temporal super-resolution network,
and the video decoder.
Our other basic settings follow the open-sourced Stable Diffusion code \footnote{https://github.com/CompVis/stable-diffusion} and remain unchanged.
All our experiments are conducted on 64 A100-80G GPUs.

%\vspace{1mm}
%\noindent\textbf{Evaluation.}
We evaluate our VideoGen on UCF-101 \cite{soomro2012dataset} 
%Sky Time-lapse \cite{xiong2018learning}, 
and MSR-VTT \cite{xu2016msr}. 
For MSR-VTT, 
we use all $59,800$ captions from the test set to calculate CLIPSIM \cite{wu2021godiva} 
(average CLIP similarity between video frames and text) following~\cite{singer2022make, wu2022nuwa}.
UCF-101 contains 13,320 video clips from 101 categories that can be grouped into body movement, human-human interaction, human-object interaction, playing musical instruments, and sports.
For UCF-101,
we follow Make-A-Video~\cite{singer2022make}
and construct the prompt text
for each class. 

Following previous methods \cite{singer2022make, ho2022video, hong2022cogvideo}, we report commonly-used Inception Score (IS) \cite{saito2020train} and Frechet Video Distance (FVD) \cite{unterthiner2018towards} \cite{unterthiner2018towards} as the evaluation metrics on UCF-101.
During the evaluation, we only generated 16$\times$256$\times$256 videos, because the C3D model \cite{tran2015learning} for IS and FVD, and the clip image encoder \footnote{https://github.com/openai/CLIP} for CLIPSIM do not expect higher resolution and frame rate.

%Since Sky Time-lapse does not have any text information, we use a BLIP2 \cite{li2023blip} model to extract the caption of the first frame of each video clip as the text prompt. 
%Similarly, for quantitative evaluation of Sky Time-lapse, we use FVD and KVD.

\subsection{Results}
\noindent\textbf{Quantitative evaluation.}
We compare our VideoGen with some recent text-to-video generation methods, including Make-A-Video \cite{singer2022make}, CogVideo \cite{hong2022cogvideo}, VDM \cite{ho2022video}, LVDM \cite{he2022latent}, TATS \cite{ge2022long}, MagicVideo \cite{zhou2022magicvideo}, DIGAN \cite{yu2022generating} and N{\"u}wa \cite{wu2022nuwa}, etc.
Because ImagenVideo \cite{ho2022imagen} has neither open source nor public datasets results, we have only made a qualitative comparison with it.
The results on MSR-VTT are given in Table \ref{table:MSR-VTT}.
We can see that
our VideoGen achieves the highest average CLIPSIM score without any fine-tuning on MSR-VTT,
proving that the generated videos and texts have good content consistency. 

The results on UCF-101 given in Table \ref{table:UCF101} show that
in the cases of both the zero-shot 
and finetuning settings, 
the IS score of VideoGen performs the best.
In the zero-shot setting,
the IS score is greatly improved compared to the second best,
from $33$ to $71.6$. 
The IS index measures the quality and category diversity of generated video
and
the high IS index indicates that the video quality and category diversity of our generated videos are excellent.
\begin{figure}[h]
	\centering	\includegraphics[width=0.48\textwidth]{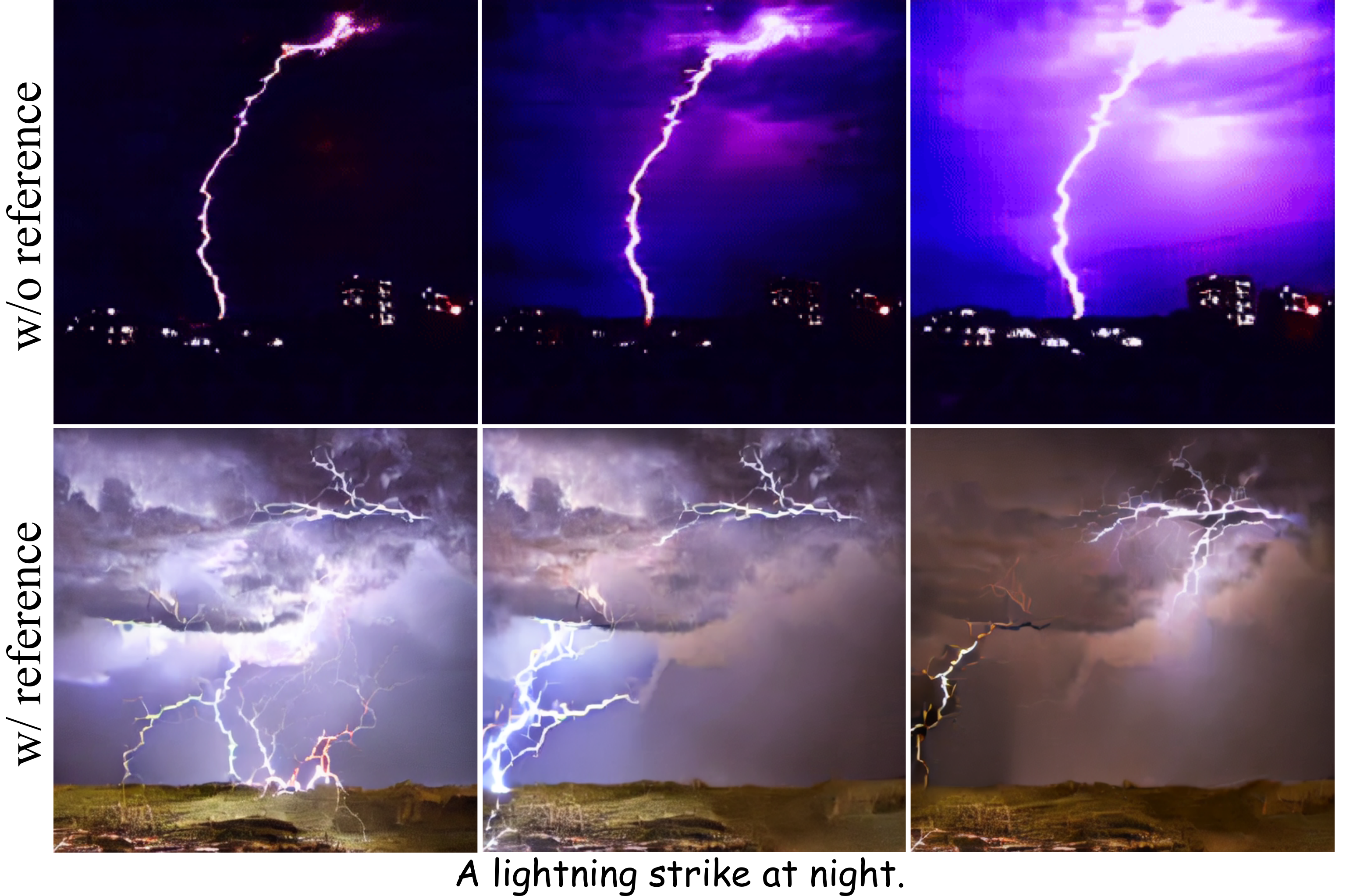}
	\caption{Visual comparison without and with the use of reference image. As we can see, the frames with reference-guided have more texture details in dark cloud and grass areas. Please zoom in to see more details.}
	\label{fig:ab_ref}
\end{figure}

The key reason for better results
from our approach
is that we generate a high-quality reference image
using a well-trained \texttt{T2I} generation model,
and accordingly the quality of generated video content is improved.

We also report the results
in terms of FVD
that measures the gap between the distribution of real videos and generated videos.
Our approach performs the second best in the zero-shot setting.
The most possible reason is that our training data distributes more differently from the UCF-101 dataset than the training data used by Make-A-Video.
In the fine-tuning setting,
we do not fine-tune the text-to-image generation model,
the flow-based temporal super-resolution model,
and the video decoder,
and only fine-tunes the first latent video diffusion model.
We guess that
our FVD score would be better
if we fine-tune the text-to-image model for generating 
a reference image whose content
matches the distribution 
of UCF-101.
The fine-tuning setting
is not our current focus,
and our current goal is general T2V generation.

\vspace{1mm}
\noindent\textbf{Qualitative evaluation.}
In Figure~\ref{fig:frameresults}, we show some examples generated from our VideoGen. 
Our results show rich and clear texture details, 
and excellent
temporal stability and motion consistency. 
In Figure~\ref{fig:ab_qualitive_com}, 
we make a visual comparison with the two recent \texttt{T2V} methods, Imagen Video \cite{ho2022imagen} and Make-A-Video \cite{singer2022make}. 
It can be seen that although the video resolution of ImagenVideo reaches 1280$\times$768, the frames are very blurry compared with our result.
Compared with Make-A-Video, 
the lake ripples, boats and trees in our video are clearer.
%More examples are available at \url{https://videogen.github.io/VideoGen/}.

% The cause is that the generated images of \texttt{T2I} are already good enough, so migrating the appearances of \texttt{T2I} results to \texttt{T2V} videos makes the generated videos more relevant and of higher quality. 

\begin{figure}[h]
	\centering	\includegraphics[width=0.48\textwidth]{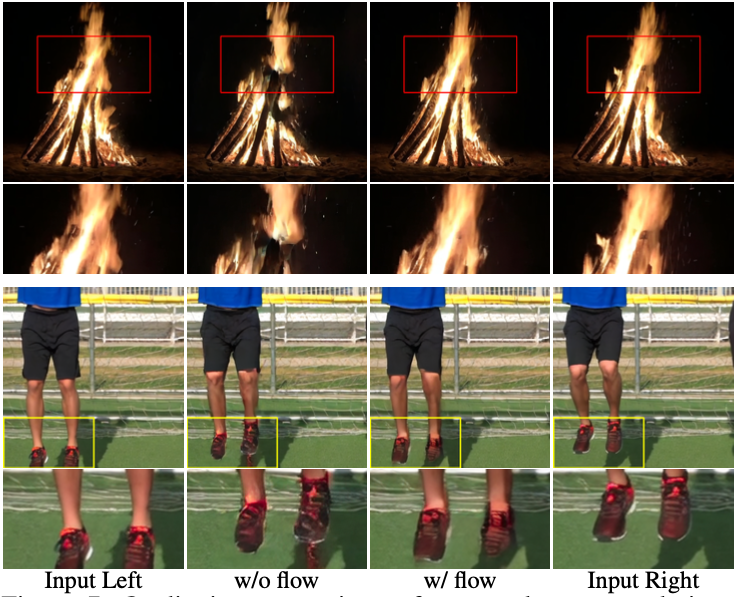}
	\caption{Qualitative comparison of 
 temporal super-resolution without and with using motion flow. 
 Using motion flow,
 the interpolated frame is more stable 
 and more consistent to input left and right frames for the top example,
 and visually better for the bottom example.
 The first and third rows are two examples,
 and the second and four rows are zoomed-in 
 of the patches in the red and yellow box.}
	\label{fig:ab_fvli}
\end{figure}

\subsection{Ablation Study}
\begin{table}[t]
\centering
\setlength{\tabcolsep}{17pt}
\renewcommand{\arraystretch}{1.3}
\footnotesize
\caption{Effect of reference guidance. 
We report average CLIPSIM score on 1000 texts 
randomly selected from the MSR-VTT testset. 
We also report the IS scores on the UCF101 dataset in the zero-shot setting.}
\label{table:ab_ref_clipsim_is}
\centering
% \resizebox{0.48\textwidth}{!}{
\begin{tabular}{l c c}
\hline
 & CLIPSIM $\uparrow$ & IS $\uparrow$ \\
\hline
without reference & 0.2534 & 26.64 $\pm$ 0.47\\

with reference & 0.3127 & 71.61 $\pm$ 0.24\\
\hline
\end{tabular}
% }
\end{table}

\noindent\textbf{Reference image
from text-to-image generation.}
In order to evaluate the effect of our \texttt{T2V} strategy guided by \texttt{T2I} reference, 
we conducted experiments
by removing the reference condition for cascaded latent diffusion models.
We randomly selected 1000 text prompts from the 59800 MSR-VTT test set and 
compared the CLIPSIM scores.
We also compared the IS index under zero-shot setting on the UCF-101 dataset. The comparison is given in Table~\ref{table:ab_ref_clipsim_is}.
One can see that
the \texttt{T2I} reference images greatly improve the IS and CLIPSIM scores. 
This empirically verifies the effectiveness of the reference image:
improving the visual fidelity and helping the latent video diffusion model
learn better motion.
Figure~\ref{fig:ab_ref} shows the visual comparison from the same text prompt.
We can see that the visual quality and the content richness with reference image are much better.
In Figure ~\ref{fig:dif_ref}, we show three different reference images, with the same text prompt, our VideoGen can generate different videos.

\noindent\textbf{Flow-based temporal super-resolution.}
We demonstrate the effectiveness of our flow-based temporal super-resolution by replacing flow-guided with spherical-interpolation guided.
The comparison with two examples
are given in Figure~\ref{fig:ab_fvli}.
We can observe that 
with motion flow the interpolated frames is more stable and continuous. 
Without flow-guided, as shown in Figure~\ref{fig:ab_fvli},
the fire is broken and the right shoe has artifacts.
% However, without warped latent code as condition, the results generated by TSR are unstable and do not preserve temporal consistency.
% %
% And for large motions in the second row, IFRNet could not capture accurate intermediate locations and TSR produces obvious artifacts. 
% %
% In summary, generating flow-guided latent codes as condition and adopting it to perform iterative latent variable refinement could achieve both high fidelity and emporal consistency.

\begin{figure}[t]
	\centering	\includegraphics[width=0.48\textwidth]{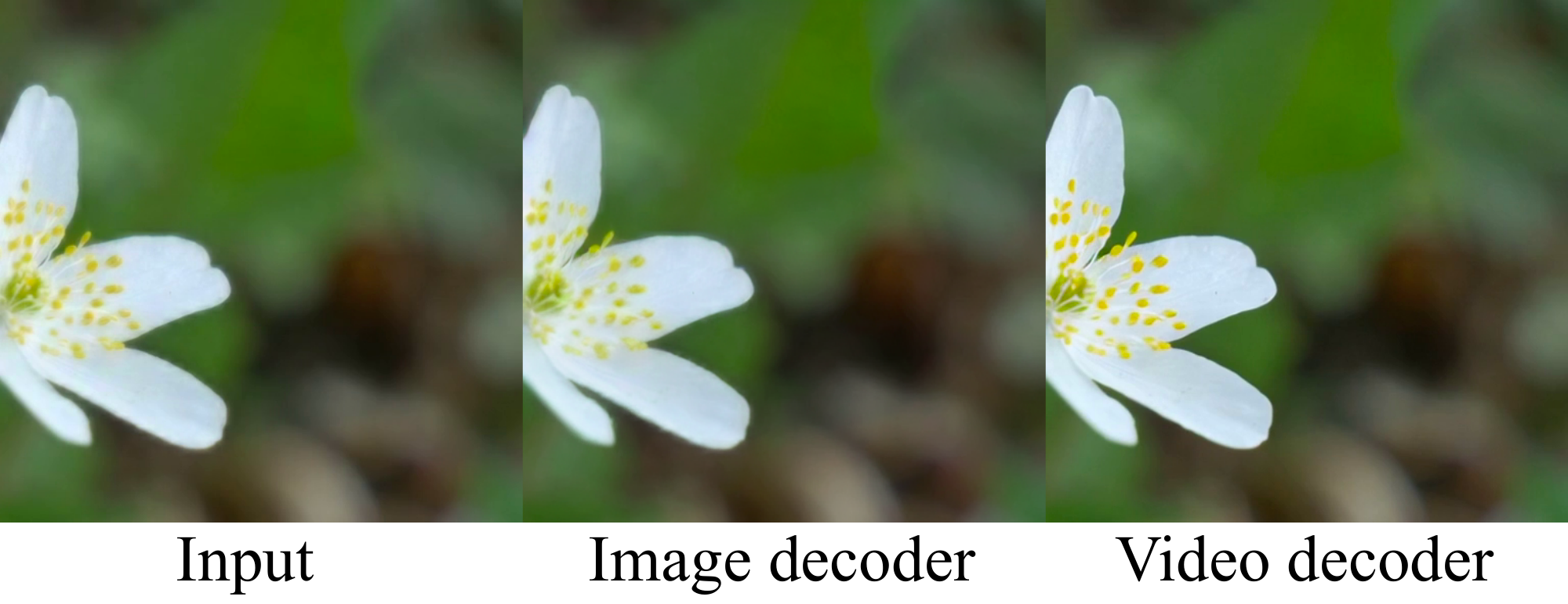}
	\caption{Visual comparison for the effectiveness of video decoder. The texture details of the the pistil and petals in our restored frame are clearer than those of original image decoder in the Stable Diffusion.}
	\label{fig:ab_vd}
\end{figure}

\noindent\textbf{Video decoder.}
Figure~\ref{fig:ab_vd} shows the visual comparison results between our video decoder and the original image decoder of the auto-encoder in Stable Diffusion. 
The frame from our video decoder has sharper textures.
This is because we perform various degradations on the inputs during training, so that our video decoder has enhanced effect.
Furthermore, the videos restored from the video decoder are temporally smoother.

\subsection{User Study}
\begin{figure}[t]
	\centering	\includegraphics[width=0.48\textwidth]{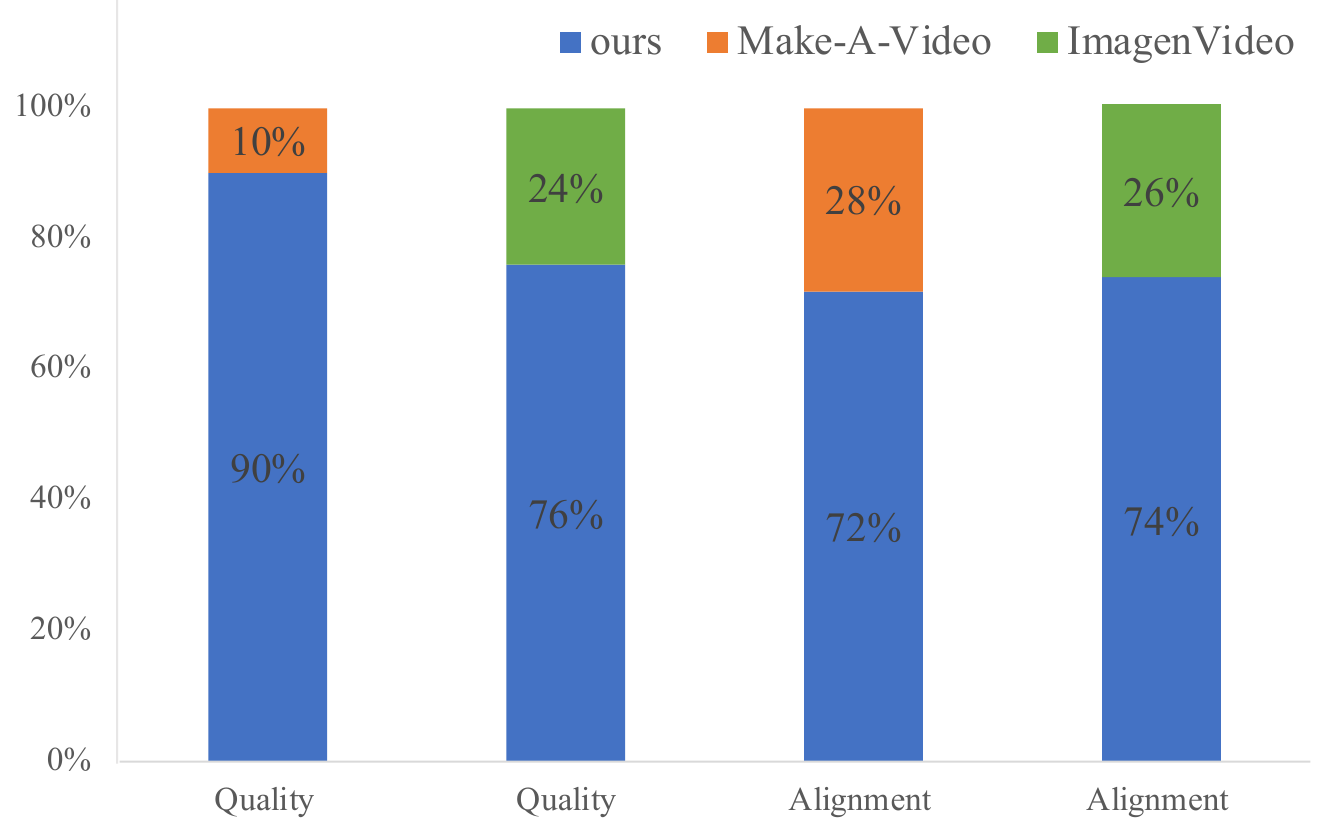}
	\caption{User Preferences. The first two bars are human evaluation results of our method compared to Make-A-Video and ImagenVideo for video quality (corresponding to the question: "Which video is of higher quality?"), respectively. Comparison with Make-A-Video, results from our approach are preferred $90\%$. Compared with ImagenVideo, $76\%$ of our options are chosen. The latter two reveal the users' preference for text-video alignment ("Which video better represents the provided text prompt?"). Similarly, our VideoGen also outperforms baseline methods by a large margin.}
	\label{fig:user_study}
\end{figure}

Because Make-A-Video ~\cite{singer2022make} and ImagenVideo ~\cite{ho2022imagen}, the two best performing methods at present, are not open sourced, we use the demos shown on their webpages for human evaluation. 
We conduct the user study on an evaluation set of 30 video prompts (randomly selected from the webpages of Make-A-Video and ImagenVideo). 
For each example, we ask 17 annotators to compare the video quality (“Which video is of higher quality?”) and the text-video content alignment (“Which video better represents the provided text prompt?”) between two videos from the baseline (ImagenVideo or Make-A-Video) and our method, presented in random order.
As shown in Figure ~\ref{fig:user_study}, in the video quality comparison with Make-A-Video, results from our VideoGen are preferred $90\%$. Compared with ImagenVideo, $76\%$ of our options are chosen.
Similarly, for the user study of the text-video alignment, our VideoGen also outperforms baseline methods by a large margin.

\section{Conclusion}
We present VideoGen, a text-to-video generation approach,
and report the state-of-the-art video generation results.
The success stems from:
(1) Leverage the SOTA text-to-image generation 
system to generate a high-quality reference image,
improving the visual fidelity of the generated video;
(2) Use the reference image as a guidance
of latent video diffusion,
allowing the diffusion model to
focus more on learning the motion;
(3) Explore high-quality unlabeled (unpaired)
video data to train a video decoder
that does not depends on video-text pairs.

% The proposed text-to-video generation approach, VideoGen, a text-to-video generation method conditioned on a text prompt and a reference image, using cascaded diffusion models processed in the latent space. In particular, we use a reference image generated by a well-trained text-to-image model as an additional condition, which makes the task less difficult and the generated videos are of higher quality and more controllable. 
% In addition, we propose a flow-guided video latent interpolation diffusion model, compared with pixel-level interpolation, which can produce arbitrary multipliers of interpolation frames without suffering from blurring. Finally, an enhanced video-level decoder is used to make the generated video more clearer and smoother. Since models are trained in the latent space, we reduce the resources required for training while still guaranteeing high-quality video generation. Extensive experiments demonstrate the superiority of our method.

\clearpage
\newpage

{\small
\bibliographystyle{ieee_fullname}
\bibliography{egbib}
}

\end{document}